# COVID-19 Probability Prediction Using Machine Learning: An Infectious Approach

Mohsen Asghari Ilani[1,*], Saba Moftakhar Tehran[2], Ashkan Kavei[3], Arian Radmehr[4]

[1,*]*School of Mechanical Engineering, College of Engineering, University of Tehran, Tehran, Iran*
[2]*School of Electrical and Computer Engineering, University of Kashan, Kashan, Iran*
[3]*Mechanical Engineering, Islamic Azad University Science and Research Branch, Tehran, Iran*
[4]*Department of Computer Engineering, Islamic Azad University, South Tehran Branch, Tehran, Iran*

## Abstract

The ongoing COVID-19 pandemic continues to pose significant challenges to global public health, despite the widespread availability of vaccines. Early detection of the disease remains paramount in curbing its transmission and mitigating its impact on public health systems. In response, this study delves into the application of advanced machine learning (ML) techniques for predicting COVID-19 infection probability. We conducted a rigorous investigation into the efficacy of various ML models, including XGBoost, LGBM, AdaBoost, Logistic Regression, Decision Tree, RandomForest, CatBoost, KNN, and Deep Neural Networks (DNN). Leveraging a dataset comprising 4000 samples, with 3200 allocated for training and 800 for testing, our experiment offers comprehensive insights into the performance of these models in COVID-19 prediction. Our findings reveal that Deep Neural Networks (DNN) emerge as the top-performing model, exhibiting superior accuracy and recall metrics. With an impressive accuracy rate of 89%, DNN demonstrates remarkable potential in early COVID-19 detection. This underscores the efficacy of deep learning approaches in leveraging complex data patterns to identify COVID-19 infections accurately. This study underscores the critical role of machine learning, particularly deep learning methodologies, in augmenting early detection efforts amidst the ongoing pandemic. The success of DNN in accurately predicting COVID-19 infection probability highlights the importance of continued research and development in leveraging advanced technologies to combat infectious diseases.

## I. INTRODUCTION

Health represents the cornerstone of any society, yet the world currently grapples with a profound health crisis due to the widespread dissemination of the coronavirus. The global COVID-19 pandemic has inflicted extensive loss of life and has profoundly impacted individuals, both directly and indirectly. Day by day, the infection and mortality rates due to the coronavirus continue to surge [1], causing widespread devastation. Additionally, the global economy reels under the weight of this epidemic, with medical infrastructure strained to accommodate the growing number of COVID-infected individuals. Despite ongoing efforts, finding a vaccine for this disease remains a formidable challenge, with countries worldwide actively engaged in research towards this end. Various strategies have been implemented by countries to mitigate the spread of the virus, underscoring the crucial need for early detection of COVID-19 infections. Physicians advise patients to undergo COVID testing if they exhibit any symptoms [2], highlighting the significance of early detection to curb further transmission. To expedite the process of patient detection, numerous technologies and strategies have been adopted. Research into advanced techniques for COVID patient detection is rapidly progressing, emphasizing the urgent need for swift decision-making technology in the medical sector. Machine learning techniques play a pivotal role in disease detection and diagnosis, offering the potential for instantaneous tracking and diagnosis of COVID patients [3]. By analyzing diverse symptoms, machine learning algorithms are employed in classifying individuals with a likelihood of infection. Biomedical data classification using machine learning has emerged as a vital area of research in recent years. Deep learning, in particular, with its multiple hidden layers, has demonstrated superior accuracy in classifying data compared to traditional methods. Building on the growing body of research in machine learning for COVID-19 analysis, this study proposes a deep neural network classifier for classifying COVID-positive patients based on a set of common symptoms [3]. This approach analyzes data on five key symptoms and leverages it to effectively distinguish infected individuals. This offers potential for self-assessment tools and aids in identifying people who might need confirmatory testing based on their reported symptoms [3,4]. The following sections delve deeper into this research: Section 2 provides a review of existing research on machine learning applications in COVID-19 analysis. Section 3 details the methodology employed in this study, outlining the specific deep neural network architecture and the chosen symptoms for analysis. Section 4 presents the classification results achieved by the model and compares them to the performance of other methods. Finally, Section 5 concludes the paper by summarizing the key findings and outlining potential future directions for this research area [5].

## II. LITERATURE REVIEW

Artificial intelligence (AI) presents a promising avenue for tackling the complex challenges brought about by the COVID-19 pandemic. However, its effectiveness is not solely determined by technological advancements but also by the expertise and creativity of those who utilize it. The ongoing crisis has shed light on some of the limitations of AI systems, particularly their inability to adapt to slight changes in context or tasks without prior design [1]. The evaluation of AI's performance in any given scenario heavily relies on the skills and judgment of its developers and users. As we grapple with the COVID-19 emergency, it becomes crucial to recognize AI as a dynamic system, with its effectiveness shaped by human intervention. Collecting updated training data is essential to address novel challenges, such as those posed by the pandemic, as it enables both human managers and AI systems to make informed decisions and adapt strategies effectively [6]. The field of biomedical data analysis has witnessed a surge in the use of machine learning and data mining techniques. These powerful tools are revolutionizing disease detection and diagnosis by leveraging complex patterns within vast datasets [7]. They analyze a diverse range of biomedical signals and imaging modalities, automating tasks previously reliant on human expertise. The COVID-19 pandemic has further highlighted the crucial role of machine learning in patient identification and classification. Deep learning, a subfield of machine learning, has proven particularly effective in processing large volumes of medical images, such as chest CT scans. These techniques help pinpoint regions infected by the virus [8]. Convolutional neural networks, a specific type of deep learning architecture, have played a key role in automating image segmentation tasks. Additionally, feature extraction methods further enhance the accuracy of models used for classifying patients [5]. Machine learning applications extend beyond just image analysis in the fight against COVID-19. These tools are playing a vital role in various aspects of managing the pandemic, including drug discovery, patient monitoring, and even epidemic prediction. Mathematical models powered by machine learning algorithms provide valuable insights into the global spread of the virus, aiding in the design of effective containment strategies [9]. Additionally, innovations such as smartphone-based patient detection systems demonstrate the potential for cost-effective and accessible solutions in healthcare. These technologies, coupled with AI-driven predictive models, contribute to early detection and intervention efforts, thereby mitigating the impact of the pandemic [10]. Govindan et al. [14] present a Mixed-Integer Linear Programming (MILP) model to improve the resilience of infectious healthcare waste management during health crises like the COVID-19 pandemic. The study utilizes a stochastic approach to manage uncertainties and includes strategies such as new collection centers and third-party logistics to enhance network robustness. Ahmadi et al. [15] devised an approach integrating U-Net with pretrained SAM models for delineating tumor areas in breast ultrasound and mammographic imagery. Notably, the U-Net framework outperformed, particularly adept at navigating complex scenarios, and surpassed the SAM model in precision, especially for tumors characterized by non-uniform contours and indistinct edges. Ghoreishi et al. [16] developed a detection framework using an Edge-Attributed Matrix and Isolation Forest algorithm to identify key deviations like hard braking in elderly drivers, enhancing road safety and cognitive health monitoring through detailed trajectory analysis. Ahmadi et al. [17] created a supervised machine learning framework that categorizes, and segments varied geographic landscapes into distinct categories such as water, grasslands, and forests, employing digital twin technology in coastal areas. Reihanifar et al. [18] introduced a Multi-Objective Genetic Programming Model (MOMGGP) for forecasting meteorological drought using the Standardized Precipitation Index in Burdur City, Turkey. The model matches the accuracy of traditional methods but with less complexity, enhancing its practical utility in drought prediction. Ahmadi et al. [19] implemented a U-Net architecture which demonstrated substantial effectiveness in identifying cracks, offering valuable perspectives on refining the U-Net framework for enhanced segmentation tasks. Moghim et al. [20] utilizes the Weather Research and Forecasting (WRF) model to predict Cyclone Sidr in Bangladesh, enhancing rainfall simulations with Bayesian regression models and a probabilistic framework for hazard assessment. Ahmadi et al. [21] presented a deeply supervised adaptable neural network that classifies Alzheimer's disease severity into four levels using MRI images, showing improved diagnostic accuracy through multitask feature extraction. Arsalani et al. [22] proposed a Mixed-Integer Programming (MIP) model for the blocking and routing shipment problem, using valid inequalities to improve accuracy and solve large-scale instances through exact optimization methods applicable across different fields. Danandeh Mehr et al. [23] developed the VMD-GP model, an innovative evolutionary method for forecasting drought in ungauged areas, combining variational mode decomposition and genetic programming to enhance prediction precision using global data. Govindan et al. [24] developed a bi-objective MILP model utilizing Industry 4.0 technologies to optimize the logistics of healthcare waste management. The model effectively combines autonomous vehicles, IoT, and RFID, enhancing efficiency and safety in handling infectious waste under uncertainty. A number of studies, such as [27-32], have focused on improving the security of data analysis and transmissions using federated machine learning in highly distributed architectures in the IoT environment, which can be promising for enhancing ML-based smart health systems in general and COVID-19 in particular.

While significant progress has been made, there is still ample room for advancement, particularly in harnessing deep learning techniques for comprehensive COVID-19 data analysis. Future research should explore additional factors beyond symptom analysis and leverage ensemble learning approaches to further enhance prediction accuracy [11]. In summary, AI holds immense promise in addressing the multifaceted challenges posed by the COVID-19 pandemic. However, its efficacy hinges on collaborative efforts between technology developers and healthcare professionals, emphasizing the importance of human expertise in navigating current and future health crises [12].

## III. Methodology

The methodology for this study involves several steps to analyze the dataset and develop a predictive model for assessing the probability of a patient being infected with COVID-19 based on their age and symptoms. Firstly, a correlation plot is generated (***Figure 1***) to examine the relationships between the features and the target variable, which is the infectious probability. Notably, the analysis reveals a significant correlation, particularly with age having a correlation coefficient of 0.36, indicating its relevance in predicting infection probability. Subsequently, a violin plot (***Figure 2***) is constructed to visualize the distribution of the features across different levels of infection status, providing insights into their varying distributions and potential predictive power.

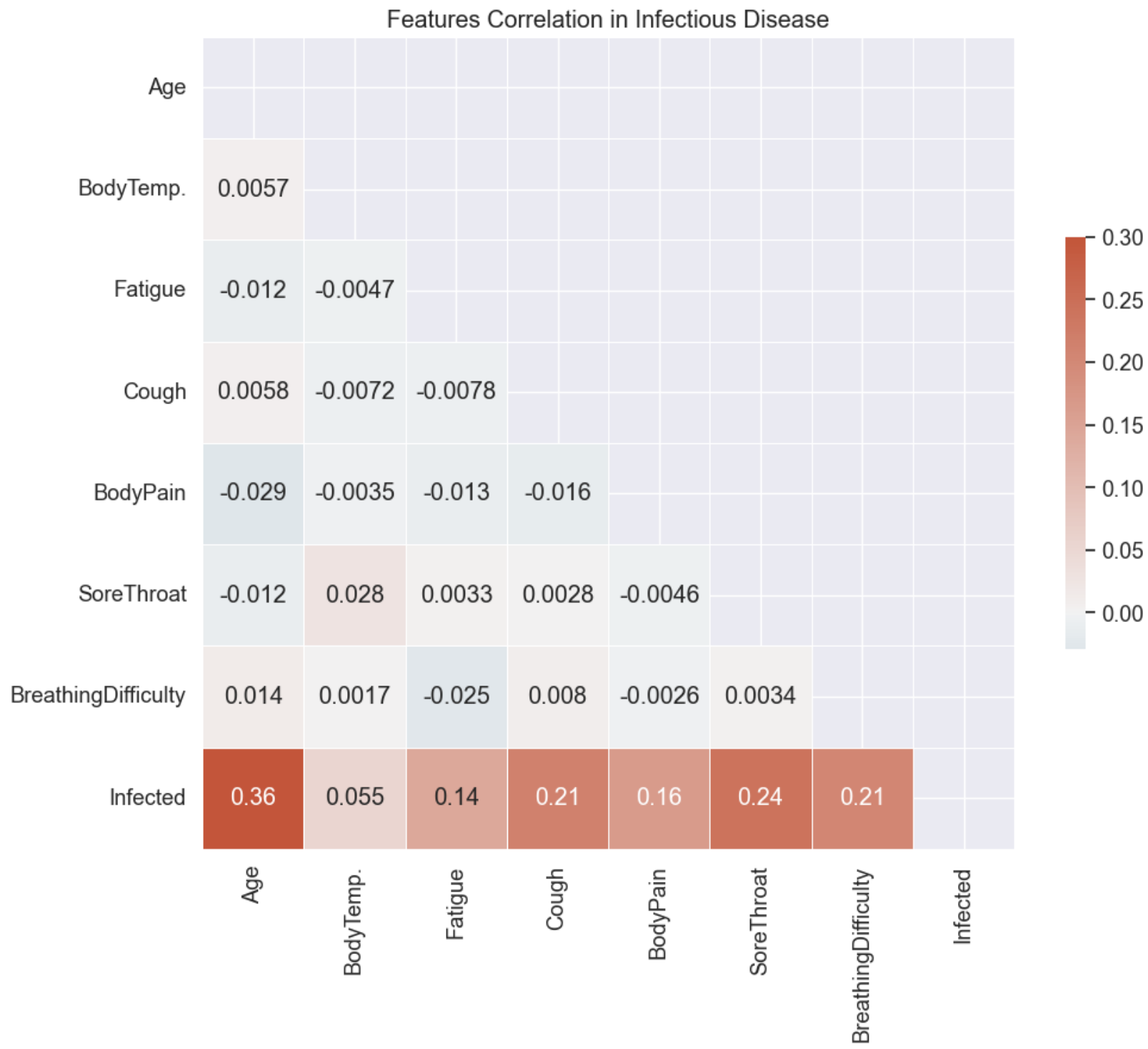

**Figure 1.** Features and Target Correlation.

The dataset consists of information regarding symptoms such as 'Fatigue', 'Cough', 'Body pain', 'Sore throat', and 'Breathing difficulty', each encoded as binary or categorical variables representing presence or severity levels. With the dataset prepared, various machine learning models are trained and tested. These models include XGBoost, LightGBM, AdaBoost, Logistic Regression, Decision Tree, Random Forest, CatBoost, KNN, and DNN. Each model is evaluated for its ability to predict the probability of infection based on age and symptoms. For example, when a new patient presents with information on age and symptoms, the trained model can predict the likelihood of the patient being infected. For instance, if a patient aged 42 reports a body temperature of 102.65°F, fatigue, body pain, and a sore throat, but no cough or breathing difficulty, the model may output a probability estimate of 46% for infection. Overall, this methodology encompasses data exploration, feature engineering, model training, and evaluation to develop a predictive framework for assessing COVID-19 infection probability based on patient characteristics and symptoms.

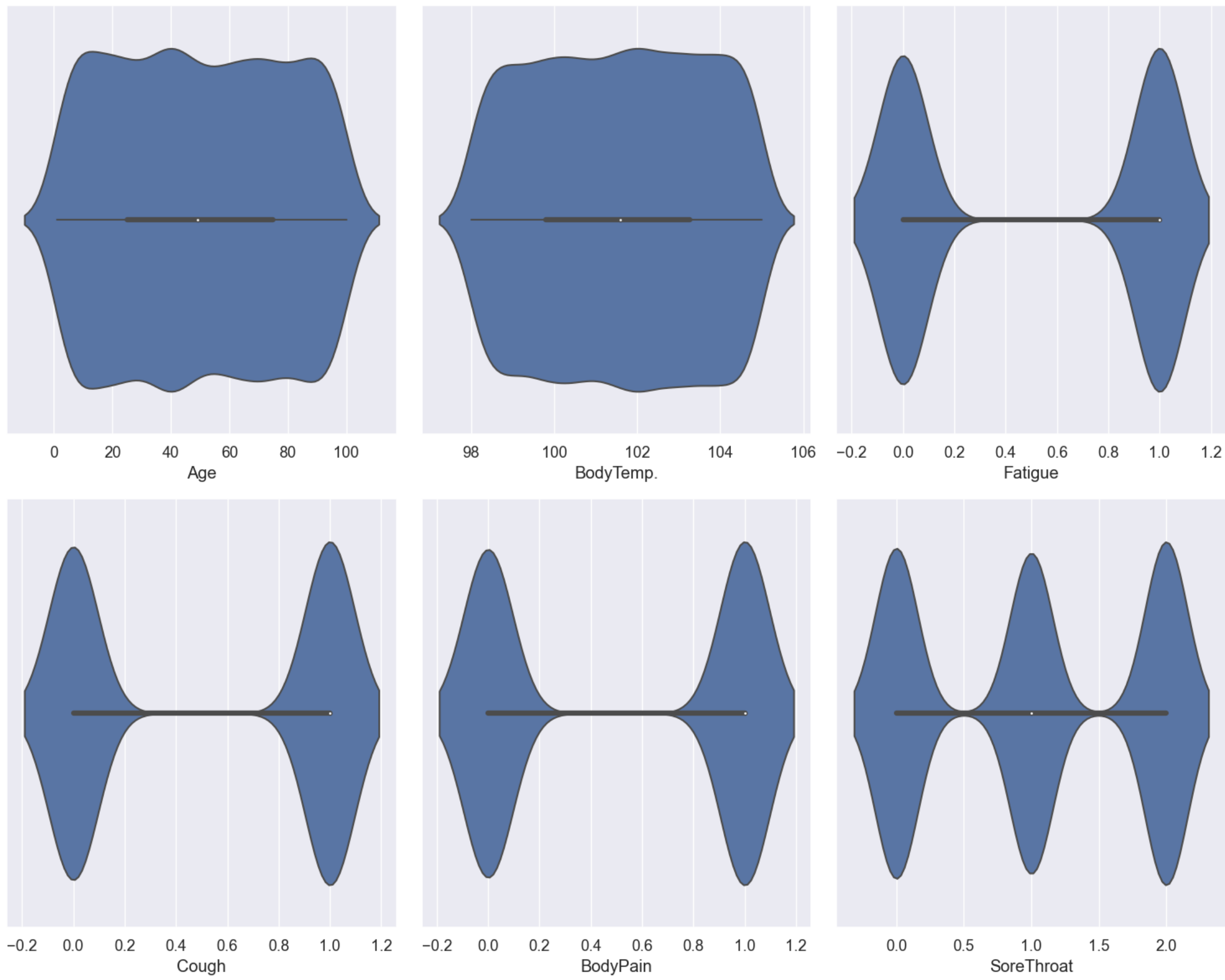

**Figure 2.** Distribution of Data by using violin plot.

### 3.1 Data Collections

The researchers utilized a data splitting approach to prepare the COVID-19 dataset for machine learning analysis. This dataset, containing 4000 instances, was divided into training and testing sets. A specific split of 80/20 was chosen, allocating 3200 instances for training the model and reserving the remaining 800 instances for testing its performance. To prevent overfitting, a common challenge in machine learning where a model performs well on training data but poorly on unseen data, the researchers implemented a k-fold cross-validation technique with k=5. This technique involves partitioning the dataset into five subsets. The model is then trained on four folds while the remaining fold is used for evaluation. This process is repeated five times, ensuring each data point is used for both training and evaluation, leading to a more robust and generalizable model. Finally, it's worth noting that the dataset itself was carefully curated, drawing from various published works and the World Health Organization (WHO). This diversity of sources helps ensure the dataset's relevance and applicability to the complexities of the COVID-19 pandemic.

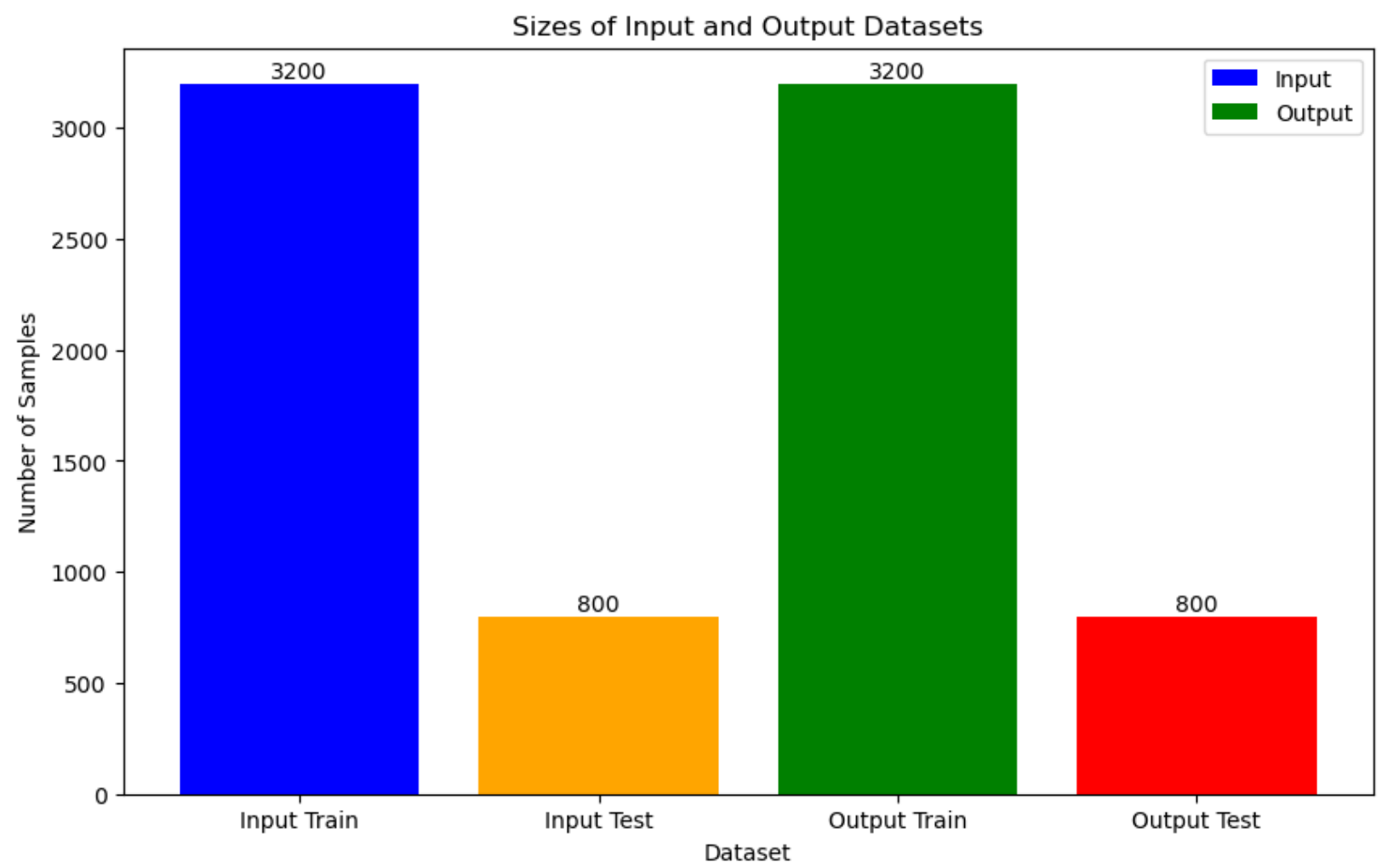

**Figure 3.** Data Splitting, training and test datasets.

## 3.2 ML Models

### 3.2.1 Linear Classification Algorithm

A classification algorithm, known as a classifier, determines its classifications based on a linear predictor function that combines a set of weights with the feature vector, as shown in the equation below:

$$y = f(\vec{w}.\vec{x}) = f(\sum_{j} w_j . x_j)$$

Here, $y$ represents the output, $f$ is the activation function, $\vec{w}$ denotes the weights, $\vec{x}$ is the feature vector, and $\vec{w}.\vec{x}$ represents the dot product of weights and features.

Given the widespread use of linear machine learning (ML) algorithms, this study focuses on employing two well-known models: logistic regression and support vector machine (SVM) with a linear kernel, also known as Support Vector Classification (SVC).

#### *3.5.1.1 Logistic Regression*

Logistic regression is a statistical method used to model the probability of a discrete outcome given an input variable [33]. While commonly applied to binary classification tasks, it can also be extended to handle multiclass scenarios. In multiclass logistic regression, where there are more than two possible outcomes, the model utilizes multiple logistic functions, each corresponding to a specific class. The equation for multiclass logistic regression can be expressed as follows:

$$P(y_i = k|x_i) = \frac{e^{w_k . x_i}}{\sum_{j=1}^{k} e^{w_j . x_i}}$$

Here, $P(y_i = k|x_i)$ represents the probability that the input sample $x_i$, belongs to class $k$, $w_k$ is the weight vector corresponding to class $k$, and $x_i$ is the feature vector of the $i$-th sample, $k$ denotes the total number of classes.

### 3.5.2 Non-Linear Classification Algorithm

However, in scenarios where classes cannot be separated by a linear boundary, non-linear classifiers become essential in using machine learning models. These classifiers are adept at handling intricate classification challenges by capturing complex patterns and relationships within data. Unlike linear models, non-linear classifiers offer enhanced performance when faced with complex datasets. Below, we present the most common models fitted to our four classes of melt pool shape, encompassing tree-based algorithms, Neural Networks (NNs) and Gaussian Naive Bayes (GaussianNB) algorithms, providing insight into the capabilities of non-linear algorithms.

#### *3.5.2.1 Tree-based model*

Tree-based models, encompassing decision trees, random forests, and gradient boosting machines, are a popular choice in machine learning for both classification (categorizing data) and regression (predicting continuous values) tasks. These models work by iteratively dividing the feature space (the set of variables used for prediction) into smaller and more specific regions. Each split is based on a chosen criterion, like Gini impurity (classification) or information gain (both), which measures how well a split separates different data points. This process builds a tree-like structure, where each terminal node (leaf) represents a final decision or prediction. Random forests and gradient boosting machines go a step further by combining multiple trees into robust ensemble models. These ensembles often outperform individual trees by leveraging the collective strength of diverse learning approaches. Overall, tree-based models are prized for their interpretability - meaning the logic behind their predictions is relatively easy to understand - flexibility in handling different data types, and their capability to capture complex relationships within the data itself.

##### 3.5.2.1.1 Decision Trees

Decision trees, a type of non-parametric supervised learning method, are widely used for both classification (categorizing data) and regression (predicting continuous values) tasks. In this study, we leverage decision trees specifically for classifying melt pool shapes into four distinct categories. A decision tree operates by iteratively dividing the input data space into smaller regions based on the values of specific features (measurable characteristics). This process aims to predict the target variable, which in this case is the melt pool shape category. At each internal node (decision point) within the tree, a single feature value is used to create a branching structure. This branching continues until a leaf node (terminal point) is reached, where a final prediction for the target variable (melt pool shape) is made.

For problems with multiple classes, like classifying melt pool shapes, decision trees extend their binary logic (splitting into two branches) to handle more outcomes. Instead of two branches, a multi-class decision tree creates multiple branches at each node, with each branch leading to a potential class label. The decision-making process involves recursively partitioning the feature space based on feature values. Each split aims to maximize the "purity" of the resulting subsets, meaning they should contain data points belonging primarily to a single class. This concept ensures the most accurate classification at each leaf node.

$$\begin{cases} class_1 & if\ Feature\ < threshold_1 \\ class_2 & if\ threshold_1 \leq Feature < threshold_2 \\ class_n & if\ threshold_{\,n-1} \leq Feature < threshold_{\,n} \end{cases}$$

##### 3.5.2.1.2 Random Forest

Random forests, a powerful ensemble learning method based on decision trees, excels in both classification and regression tasks. It has been widely used for classification and regression purposes in various diseases especially neurocognitive diseases [25]. During training, a random forest constructs a multitude of individual decision trees. For classification problems, the final prediction is determined by the mode (most frequent class) among the predictions from all trees in the forest. Each tree within the forest is trained on a unique combination of two random subsets: a subset of the training data and a subset of the available features. This element of randomness helps to "decorrelate" the trees, meaning they learn slightly different aspects of the data. This reduces the risk of overfitting (performing well on training data but poorly on unseen data) and leads to a more robust overall model. In our classification tasks, the random forest aggregates the predictions of all individual trees to determine the final class label for a new data point. This ensemble approach leverages the strengths of each tree, leading to improved accuracy and generalizability. Random forests are particularly well-suited for handling high-dimensional data (data with many features) and are less susceptible to the influence of noisy features or outliers compared to a single decision tree. Additionally, their ability to capture complex relationships within the data and their resistance to overfitting make them a popular choice for a wide range of machine learning applications [26].

#### 3.5.2.1.3 Gradient Boosting Trees (GBT)

Gradient boosting trees (GBT) is a powerful tree-based machine learning algorithm designed for both classification and regression tasks. It achieves high accuracy by iteratively building an ensemble of decision trees. In each step, a new tree is added to the ensemble, focusing on improving the model's performance by learning from the errors (residuals) of the previous trees. These residuals represent the difference between the actual values and the predictions made by the existing ensemble. By focusing on these errors, the new tree refines the overall model's predictions, leading to increased accuracy. GBT excels at handling complex relationships within data due to its sequential learning approach. It is particularly adept at tackling challenging data sets, including those with high dimensionality (many features) or containing noise and outliers. In such scenarios, GBT's ability to learn iteratively and focus on correcting past errors proves advantageous compared to other machine learning algorithms. This makes GBT a robust and reliable choice for various machine learning tasks where accurate predictions are critical.

##### 3.5.2.1.3.1 Extreme Gradient Boosting Machine (XGBM)

Extreme Gradient Boosting Machine (XGBoost) is an advanced implementation of the gradient boosting algorithm that has gained popularity for its speed and performance. It is designed to optimize the gradient boosting process, incorporating features such as parallel computing, regularization, and tree pruning to enhance accuracy and efficiency.

##### 3.5.2.1.3.2 Light Gradient Boosting Machine (LGBM)

The Light Gradient Boosting Machine (LGBM) emerges as a powerful gradient boosting framework, similar to XGBoost, lauded for its exceptional speed, efficiency, and scalability. This framework shines when dealing with large-scale datasets, making it a popular choice across various machine learning applications due to its consistently impressive performance. LGBM is particularly celebrated for its remarkably fast training speeds and overall efficiency. These advantages can be attributed to its unique approach to tree growth within the ensemble. LGBM utilizes a "leaf-wise" strategy, focusing on splitting the most informative leaf node at each iteration. This contrasts with XGBoost's default "depth-wise" strategy, which grows trees level by level. The leaf-wise approach often leads to faster training, especially when dealing with large datasets. Additionally, LGBM incorporates optimization techniques like Gradient-Based One-Side Sampling (GOSS) and Exclusive Feature Bundling (EFB), further contributing to its efficiency. These optimizations allow LGBM to prioritize informative data points and reduce computational costs, making it a compelling choice for large-scale machine learning tasks.

#### 3.5.2.1.3.3 Adaptive Gradient Boosting Machine (AdaBoost)

AdaBoost, short for Adaptive Boosting, is an ensemble learning method that builds a strong classifier by combining multiple weak classifiers. It works by sequentially training a series of weak learners on weighted versions of the training data. In each iteration, the algorithm focuses more on the instances that were misclassified in the previous iteration, effectively adjusting its approach to improve performance. AdaBoost assigns a weight $\alpha_t$ to each weak learner $h_t(x)$, where $t$ represents the iteration number. The final prediction $H(x)$ is then obtained by summing the weighted predictions of all weak learners:

$$H(x) = sign(\sum_{t=1}^{T} \alpha_t h_t(x))$$

Here, $T$ denotes the total number of weak learners. The sign function ensures that the final prediction is either +1 or -1, depending on the overall weighted sum of the weak learners' predictions.

#### 3.5.2.1.3.4 Categorical Gradient Boosting Machine (CatBoost)

The Categorical Gradient Boosting Machine (CatBoost) stands out as a gradient boosting algorithm specifically designed to handle categorical features effectively. Unlike other algorithms like XGBoost and LightGBM, CatBoost doesn't require pre-processing steps like one-hot encoding for categorical data. This is achieved through a unique approach called "ordered boosting." In ordered boosting, CatBoost optimizes the sequence in which trees are added to the ensemble, leading to improved performance on tasks involving categorical data.

As with other gradient boosting algorithms, CatBoost works by iteratively training decision trees on the dataset. Each tree focuses on minimizing a specific loss function, which measures the difference between predicted and actual values. This iterative process allows the model to learn from its errors and progressively improve its predictions. CatBoost's ability to handle categorical features natively, combined with its ordered boosting approach, makes it a valuable tool for various machine learning tasks when the data includes categorical variables.

### *3.5.2.2 Neural Networks (NNs)*

Neural Networks (NNs) are versatile machine learning algorithms suitable for both regression and classification tasks [13]. Within these networks, individual neurons perform linear and nonlinear transformations on input data, producing outputs that are adjusted through the iterative process of backpropagation, wherein weights and biases are updated to optimize model performance.

### *3.5.3.1 Multilayer Perceptrons (MLPs)*

Multilayer Perceptrons (MLPs) are a fundamental type of artificial neural network known for their layered architecture. These networks consist of interconnected processing units called neurons, organized into layers. A typical MLP comprises an input layer that receives data, one or more hidden layers that perform complex computations, and an output layer that generates the final prediction. Within the network, each neuron receives input signals from neurons in the previous layer. These signals are weighted, meaning their influence is adjusted based on their importance. The neuron then applies an activation function to this weighted sum, introducing non-linearity into the network's behavior. This non-linearity allows MLPs to capture complex relationships within the data that would be difficult to model with linear methods.

Thanks to their ability to learn intricate data patterns, MLPs find extensive applications in various machine learning tasks. They excel at classification (categorizing data), regression (predicting continuous values), and pattern recognition, making them a versatile tool for a wide range of problems. The specific mathematical equation governing the behavior of a single node in an MLP is quite complex, but it essentially represents the calculation performed by each neuron, involving weighted sums of inputs and an activation function to generate its output.

$$a_i = f\left(\sum_{j=1}^{n} w_{ij}.x_j + b_i\right)$$

Where, $a_i$ is the output of the $i$-th node in the layer, $f$ is the activation function applied element-wise, $w_{ij}$ is the weight connecting the $j$-th input to the $i$-th node, $x_j$ is the $j$-th input to the node, $b_i$ is the $i$-th bias term for the node, and $n$ is the number of inputs to the node.

#### *3.5.2.3 Instance-based learning algorithm (Lazy Learner)*

Instance-based learning, also known as lazy learning, stands in contrast to more traditional machine learning approaches. Unlike methods that build models during training, lazy learning algorithms defer this process until a prediction is needed. This approach involves storing the entire training dataset rather than building a generalized model. When a new data point requires classification or prediction, the algorithm retrieves the most similar instances (neighbors) from the stored training data. These neighbors are then used to make a prediction for the new instance.

K-Nearest Neighbors (k-NN) is one of the most popular instance-based learning algorithms. In k-NN, the classification of a new instance is determined by the class labels of its k nearest neighbors within the training data. The value of k, which represents the number of neighbors considered, is a crucial parameter that can be tuned to optimize the algorithm's performance. While other instance-based learning algorithms exist, such as locally weighted learning (LWL) and Case-Based Reasoning (CBR), this study focused on utilizing the k-NN model for its effectiveness and relative simplicity.

##### 3.5.2.3.1 k-Nearest Neighbors (k-NN) algorithm

The k-Nearest Neighbors (k-NN) algorithm is a versatile tool in machine learning, applicable to both classification (categorizing data) and regression (predicting continuous values) tasks. It falls under the category of instance-based learning, where predictions rely on comparisons to stored data rather than explicit model building. In k-NN, the classification or prediction for a new data point is made by considering its k nearest neighbors in the feature space (the space defined by all the features used for prediction). The feature space allows for comparisons based on relevant characteristics. For classification tasks, the k-NN algorithm assigns the new data point the majority class label of its k nearest neighbors. In regression tasks, the prediction is determined by the average value of the k nearest neighbors. The chosen value of k significantly impacts the model's performance and needs to be carefully tuned for optimal results.

## IV. Results and discussion

In the context of medical research, the integration of machine learning (ML) models represents a transformative approach to overcoming the inherent limitations of traditional clinical and practical investigations. By harnessing the power of ML algorithms, researchers can expedite processes and achieve more efficient outcomes, particularly in critical areas such as COVID-19 detection, where timely intervention is paramount for public health.

A crucial aspect of evaluating machine learning models is their ability to avoid overfitting. Overfitting is a common pitfall where a model becomes too attuned to the specific details and noise present in the training data. This can lead to misleadingly high performance on the training data itself, but the model often fails to generalize well to new, unseen data. In essence, the model learns to perfectly mimic the training data, but it does not learn the underlying patterns that truly govern the data. This results in poor performance in real-world applications where the model encounters new data not included in the training set.

To address overfitting effectively, researchers must implement robust regularization techniques and hyperparameter tuning strategies. Regularization methods such as L1 and L2 regularization, dropout, and early stopping serve to constrain the model's complexity, preventing it from fitting noise in the training data excessively. Additionally, hyperparameter tuning facilitates the optimization of model parameters to strike a balance between bias and variance, ultimately enhancing model generalizability. Moreover, the adoption of cross-validation techniques, such as k-fold cross-validation, enables researchers to assess model performance across multiple subsets of the data, providing a more comprehensive evaluation of predictive capabilities. By rigorously monitoring and mitigating overfitting, researchers can bolster the credibility and utility of ML-based COVID-19 detection systems, ensuring their efficacy in real-world deployment scenarios. To tackle the challenge of overfitting, our study employs a range of techniques, including the adjustment of hyperparameters such as learning rate and minimum child weight. Specifically, we focus on evaluating the behavior of XGBoost and LGBM models under various parameter settings to mitigate overfitting. In our investigation, we applied XGBoost and LGBM models and closely monitored their performance under different parameter configurations. ***Figure 4*** illustrates the impact of modifying the minimum child weight parameter on the XGBoost model, revealing a notable reduction in overfitting as the parameter value increases. This observation underscores the efficacy of adjusting this parameter in mitigating overfitting tendencies within the XGBoost model architecture.

However, our analysis unveils contrasting outcomes with the LGBM model, where adjusting the minimum child weight parameter did not effectively address overfitting concerns (***Figure 5***). This discrepancy highlights the nuanced behavior of different ML algorithms and emphasizes the necessity for tailored strategies to combat overfitting based on the specific characteristics of each model.

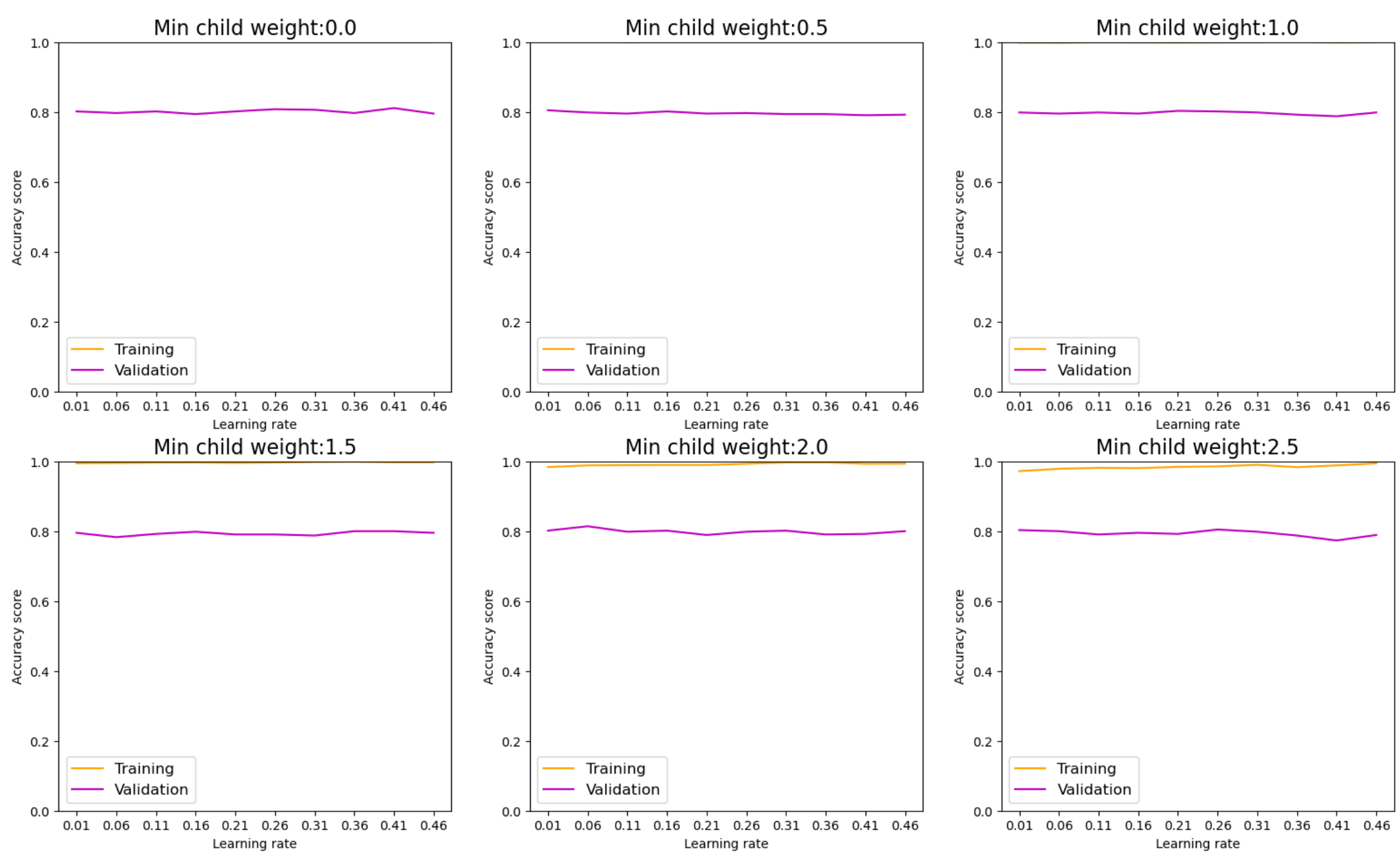

**Figure 4.** Overfit Monitoring under Learning Rate and Min Child weight in XGBoost Model.

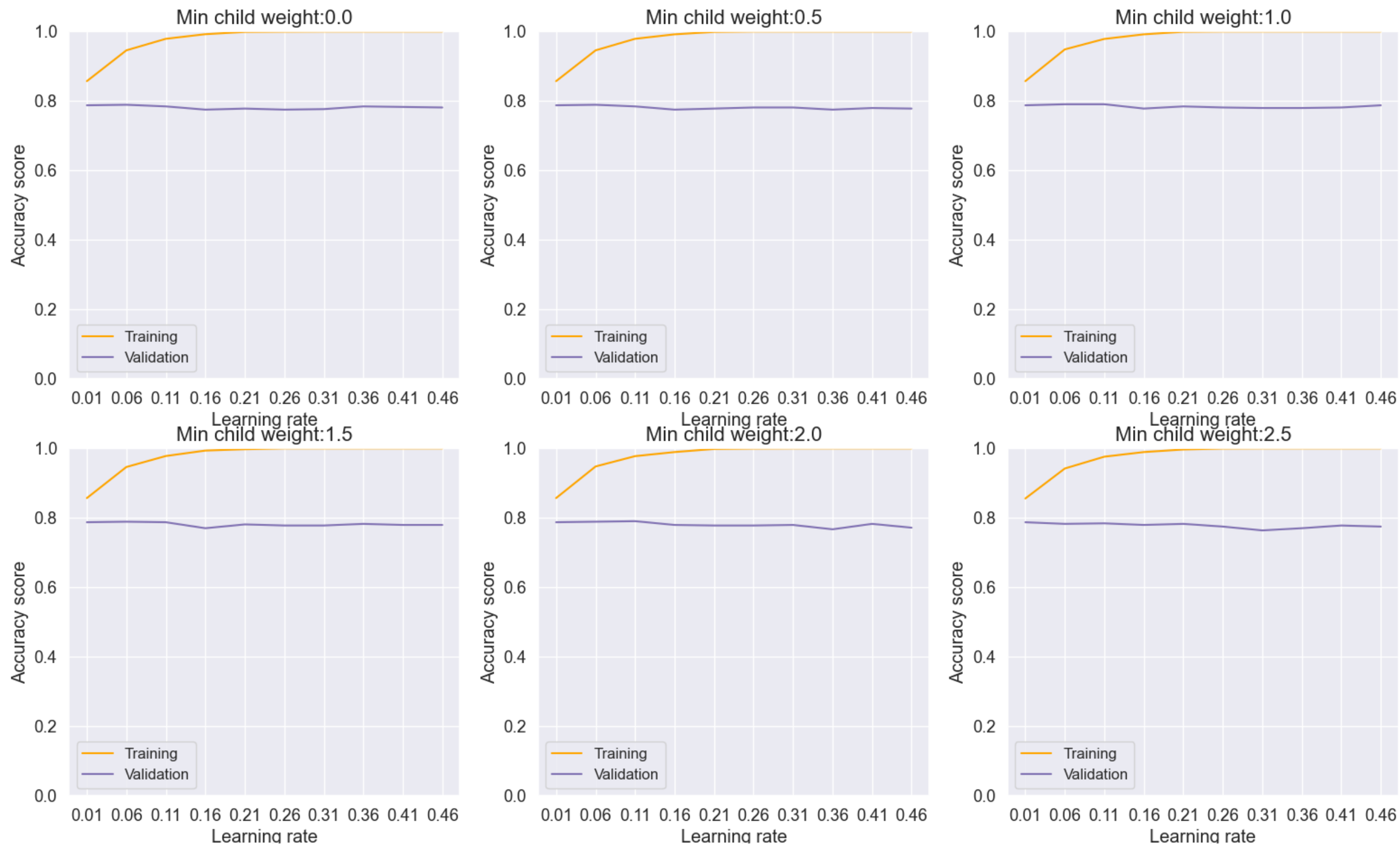

**Figure 5.** Overfit Monitoring under Learning Rate and Min Child weight in LGBM Model.

Remarkably, our analysis reveals that certain ML models exhibit consistent performance across both training and validation sets, as depicted in ***Figure 6***, ***Figure 7***, ***Figure 8***, ***Figure 9***, ***Figure 10***, and ***Figure 11***. This consistency implies minimal or negligible overfitting, underscoring either inherent mechanisms within these algorithms to prevent overfitting or favorable dataset characteristics conducive to robust model training. The observed consistency in performance across training and validation sets offers valuable insights into the underlying behavior of these ML models. It suggests that these algorithms possess inherent regularization mechanisms or are inherently resilient to overfitting tendencies. Alternatively, it could indicate that the dataset utilized in our study exhibits characteristics that facilitate effective model training without succumbing to overfitting.

These findings have significant implications for the development and deployment of ML-based systems, particularly in critical domains such as COVID-19 detection. ML models that demonstrate consistent performance between training and validation sets are more likely to generalize well to unseen data, thereby enhancing their reliability and applicability in real-world scenarios. Moving forward, further investigation into the mechanisms underlying the observed consistency in model performance is warranted. By elucidating the factors contributing to robust model training, researchers can inform the development of more resilient ML algorithms and refine model training strategies to optimize performance across diverse datasets and application domains. Furthermore, the complexity of features and the variability in individual responses to COVID-19 symptoms present additional challenges for ML models. Despite these complexities, our analysis identifies the deep neural network (DNN) model as a frontrunner, showcasing high accuracy rates in detecting infected patients. Figure 12 vividly illustrates the superior performance of the DNN model, particularly in test evaluation, underscoring its effectiveness in navigating intricate feature linkages.

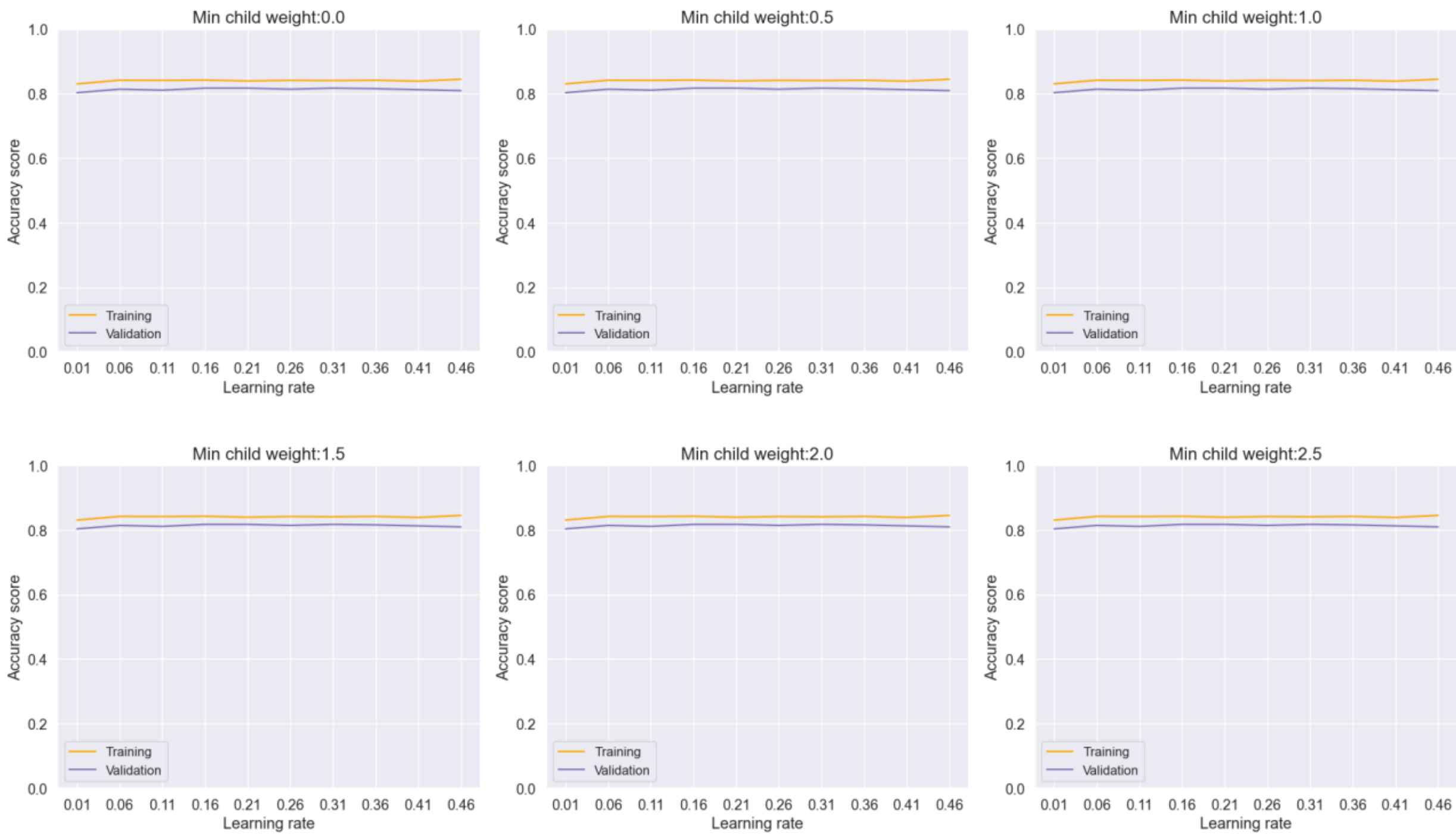

**Figure 6**. Overfit Monitoring under Learning Rate and Min Child weight in AdaBoost Model.

The remarkable performance of the DNN model in handling the complexity of COVID-19 symptomatology is noteworthy. Its ability to discern subtle patterns within the data and make accurate predictions speaks to the power of deep learning methodologies in healthcare applications. The success of the DNN model in our analysis suggests promising avenues for leveraging advanced ML techniques to address the multifaceted challenges posed by COVID-19 detection. By harnessing the capabilities of DNNs to extract meaningful insights from complex and heterogeneous data, researchers can enhance the accuracy and reliability of diagnostic systems, ultimately facilitating more effective patient care and public health interventions.

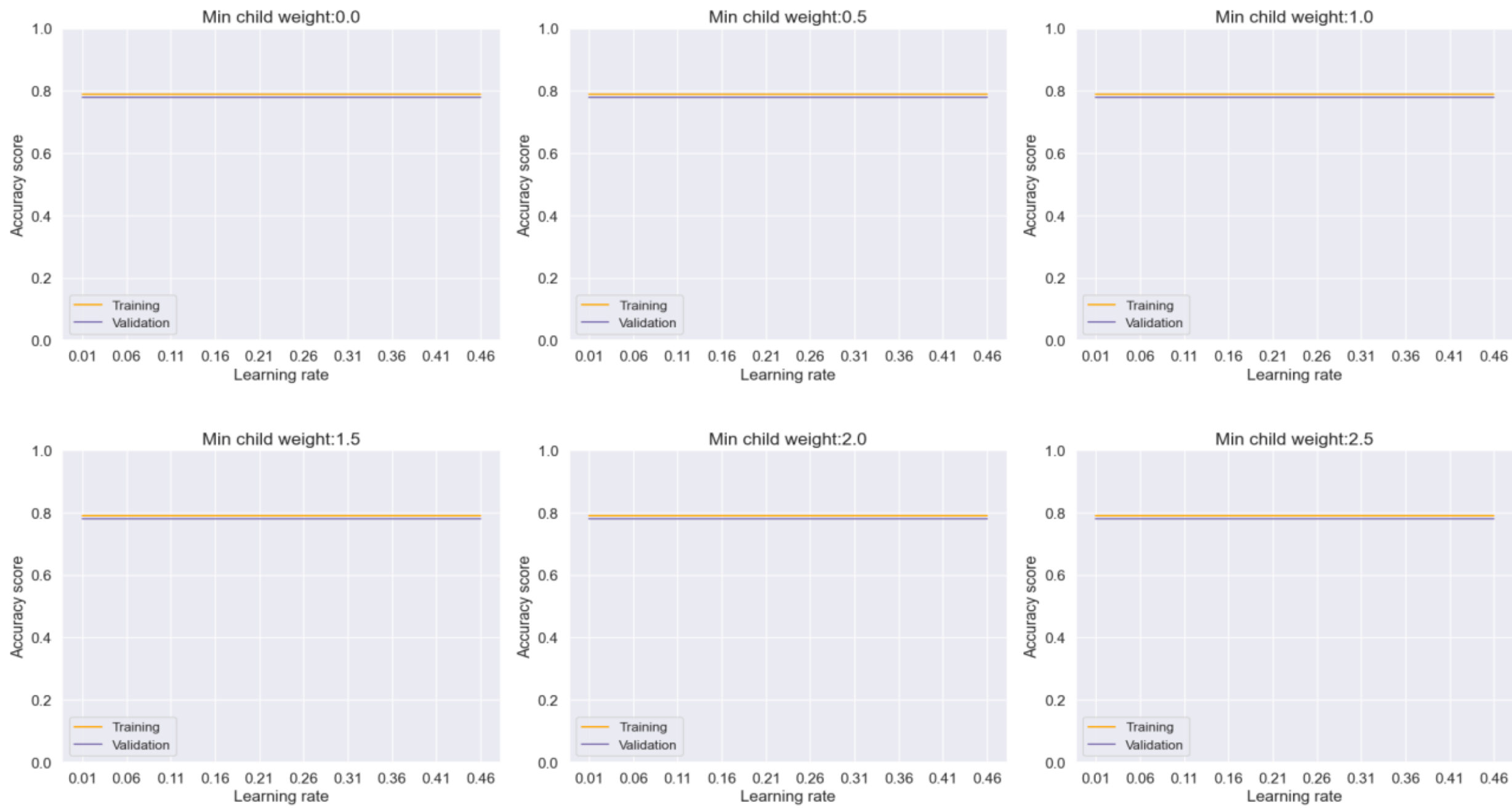

**Figure 7.** Overfit Monitoring under Learning Rate and Min Child weight in Logistic Regression Model.

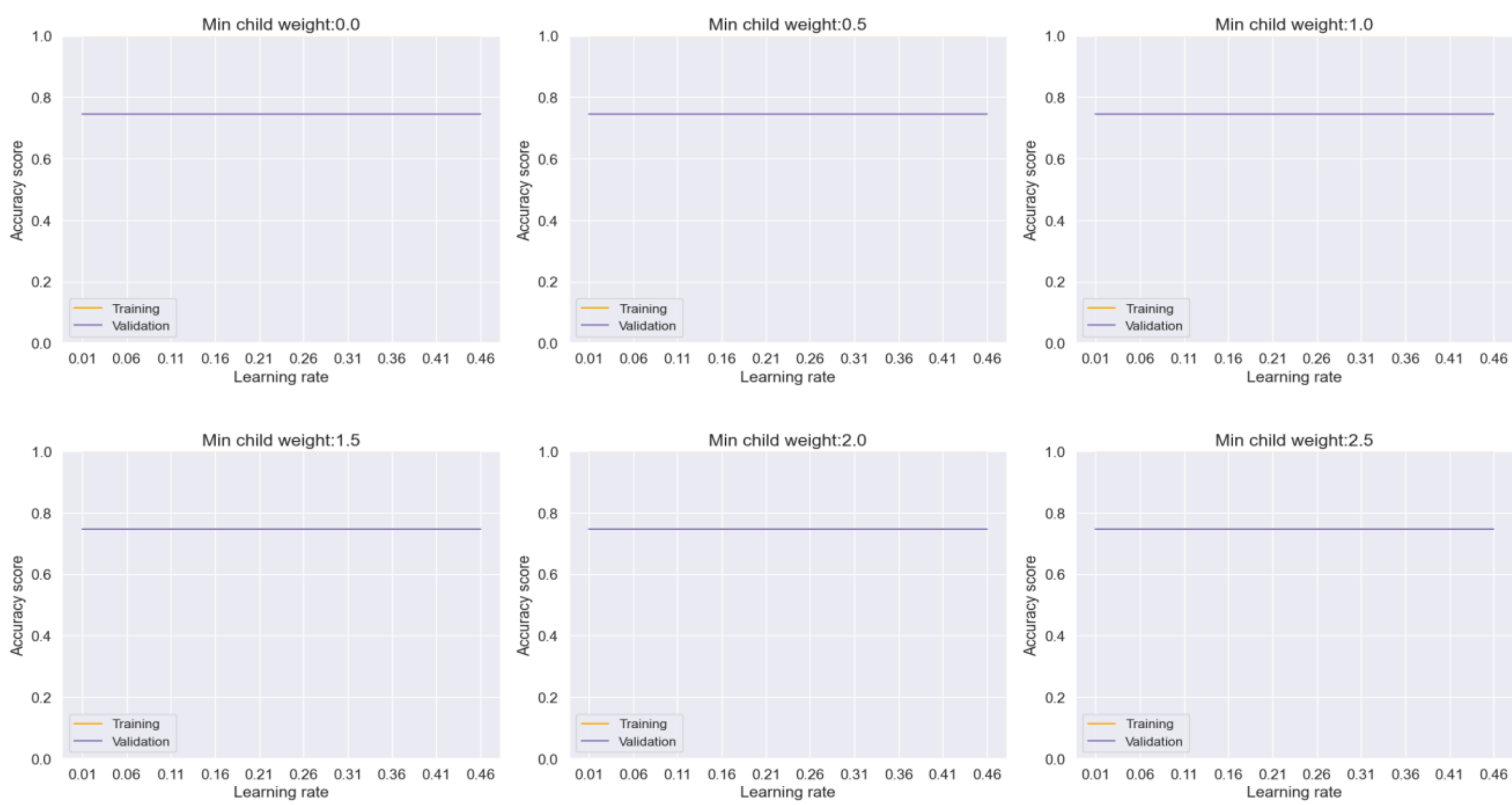

**Figure 8.** Overfit Monitoring under Learning Rate and Min Child weight in Decision Tree Model.

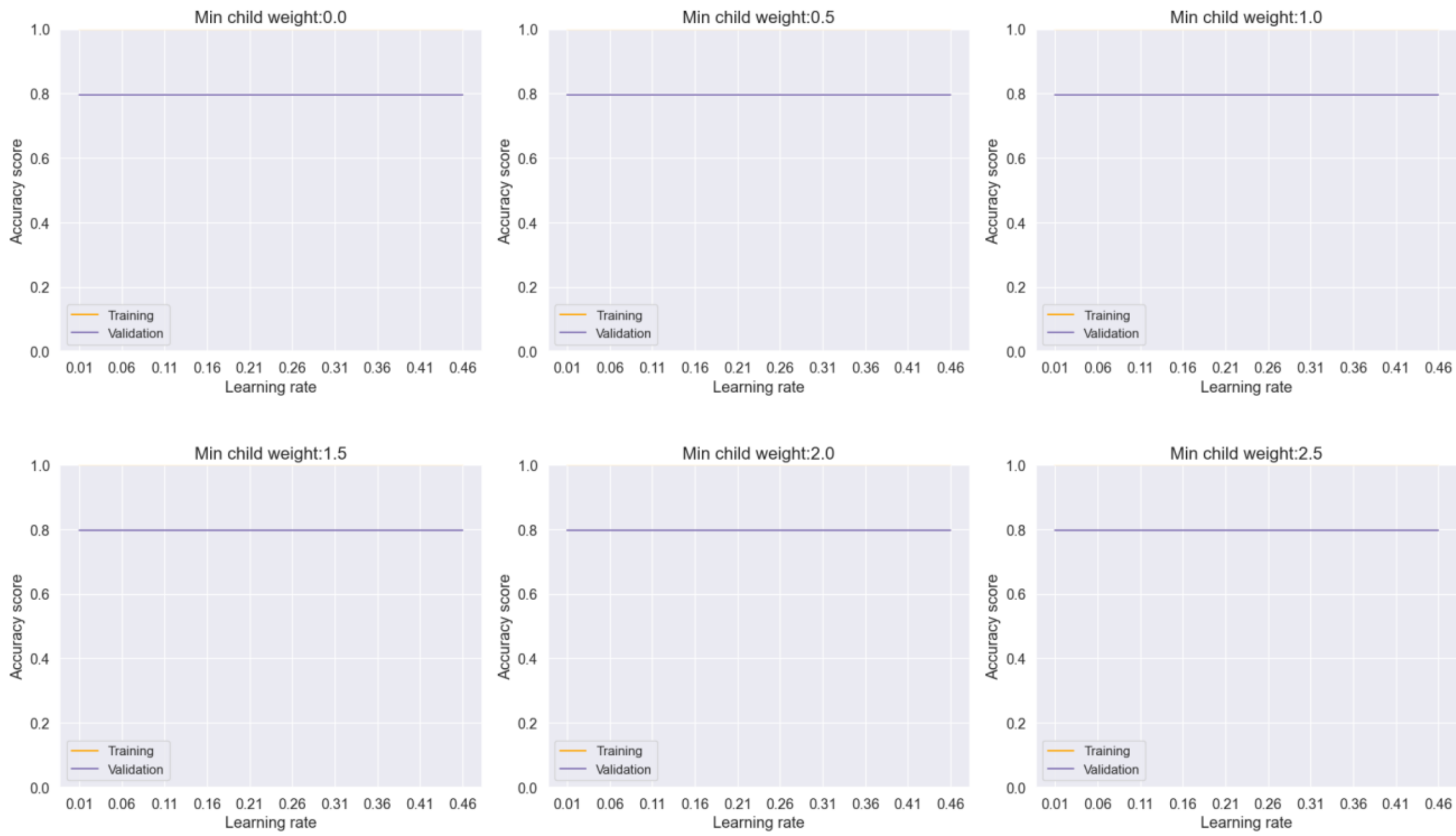

**Figure 9.** Overfit Monitoring under Learning Rate and Min Child weight in Random Forest Model.

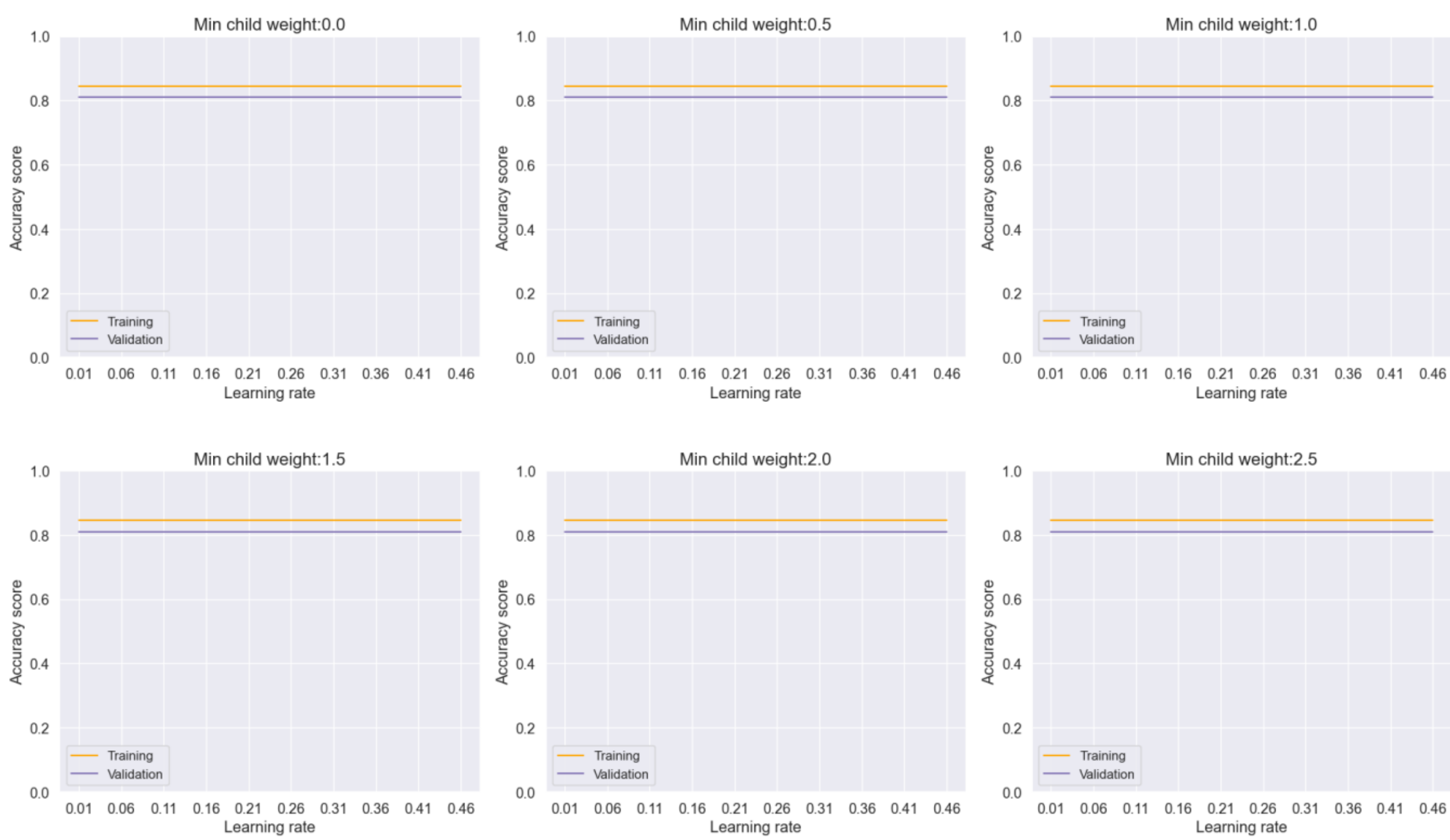

**Figure 10.** Overfit Monitoring under Learning Rate and Min Child weight in CatBoost Model.

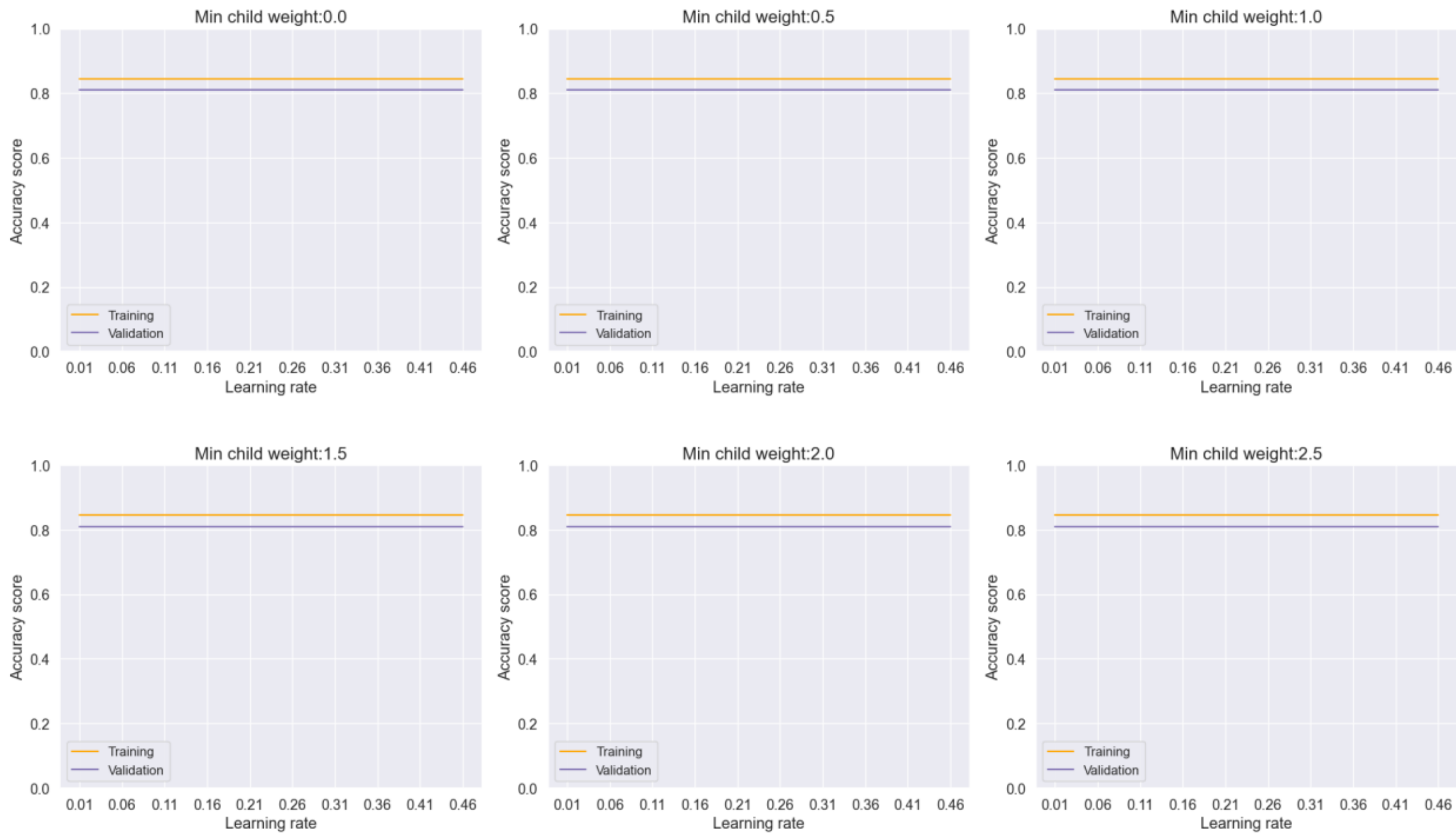

**Figure 11.** Overfit Monitoring under Learning Rate and Min Child weight in k-NN Model.

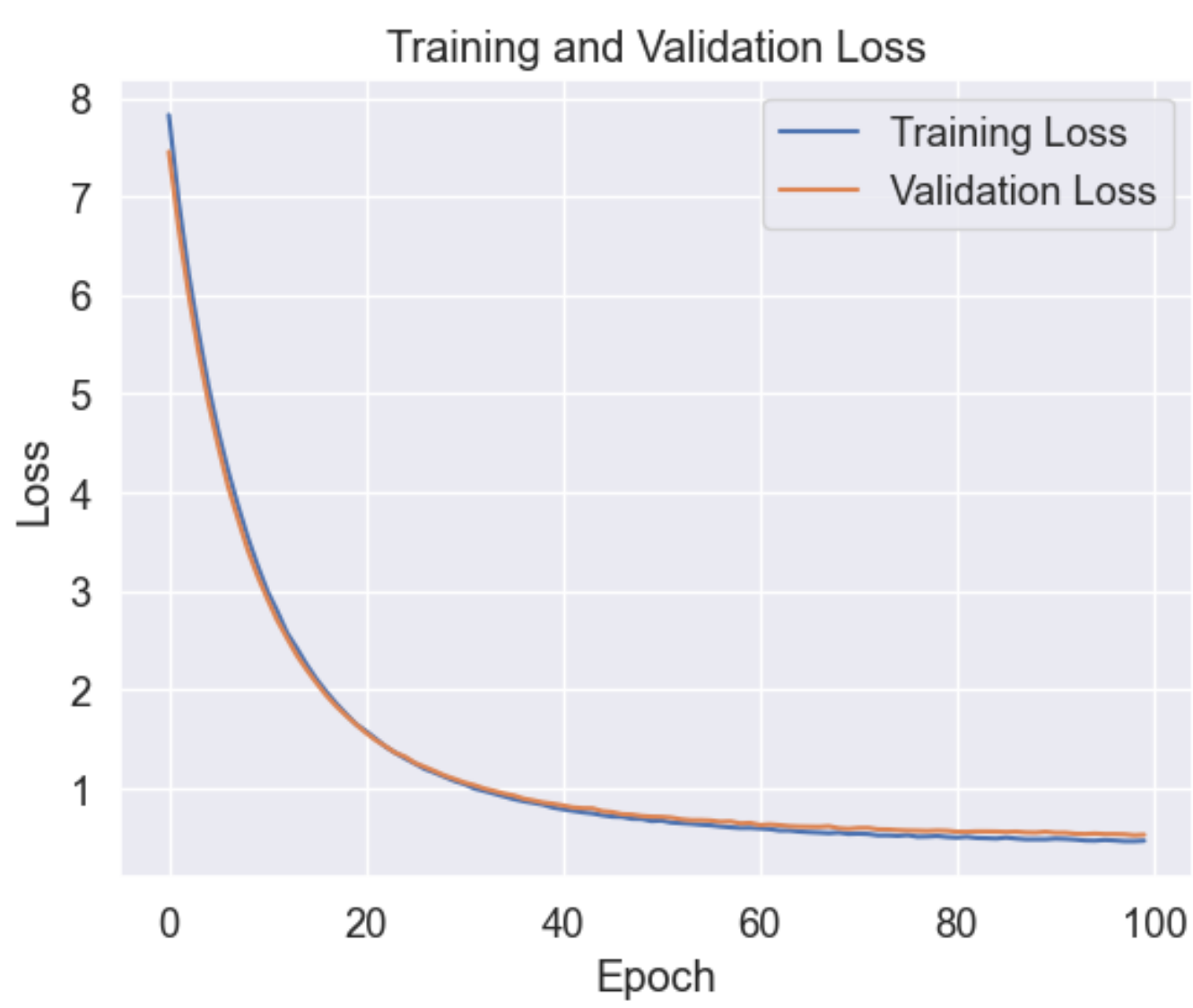

**Figure 12.** Training vs. Validation of DNN Model.

Additional validation of model performance is presented in ***Figure 13**, **Figure 14**, **Figure 15***, providing comprehensive comparisons across multiple evaluation metrics, including accuracy, precision, recall, and F1 score. These metrics serve as essential benchmarks, offering valuable insights into the overall efficacy of each ML model and elucidating the strengths and weaknesses of different approaches. The detailed analysis presented in these figures underscores the importance of evaluating ML models across a spectrum of performance metrics to gain a holistic understanding of their capabilities. By considering metrics beyond accuracy alone, such as precision, recall, and F1 score, researchers can assess various aspects of model performance, including its ability to minimize false positives, capture true positives, and maintain a balance between precision and recall.

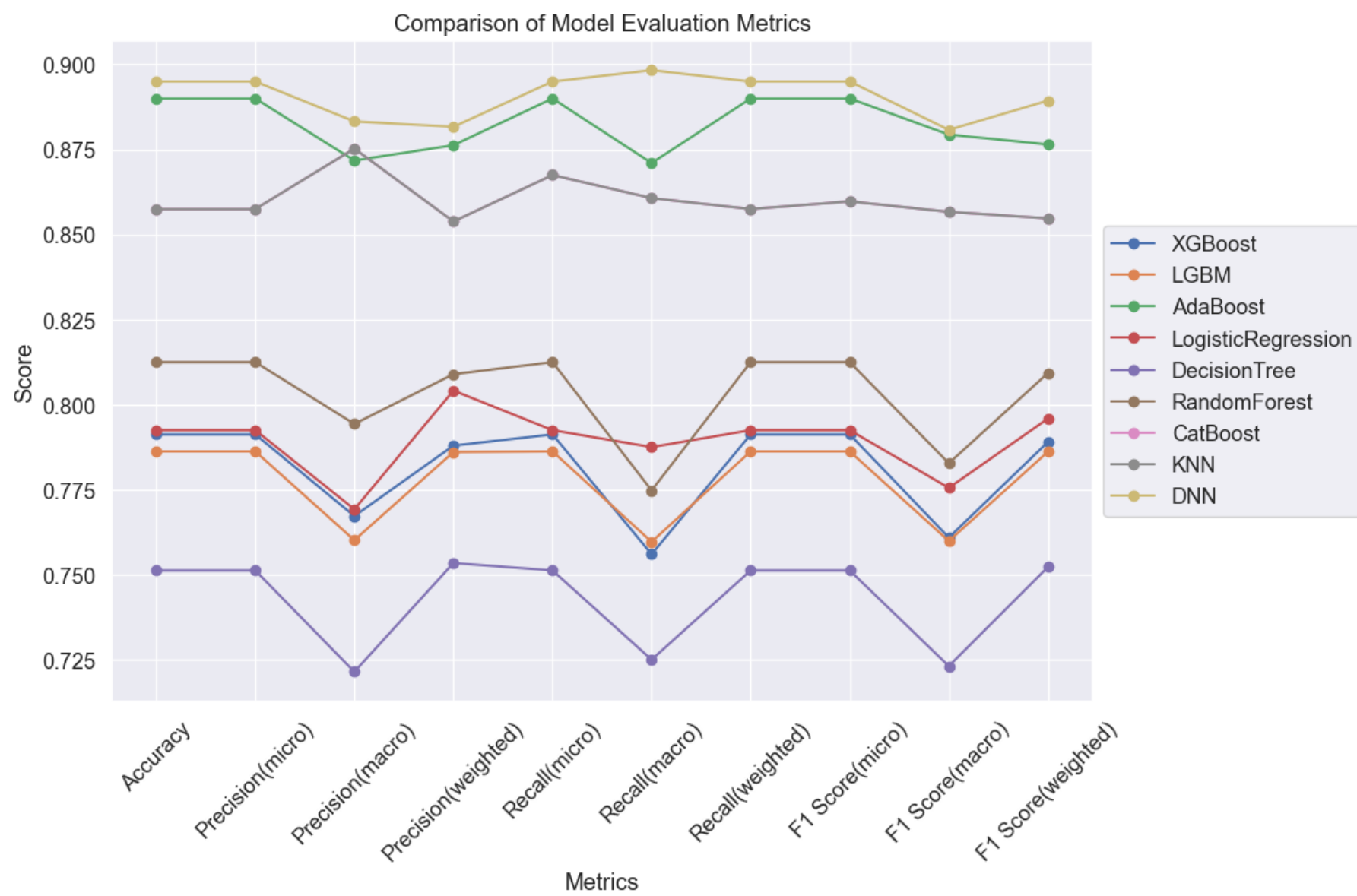

**Figure 13.** Comparison of ML Models Under the Metrics.

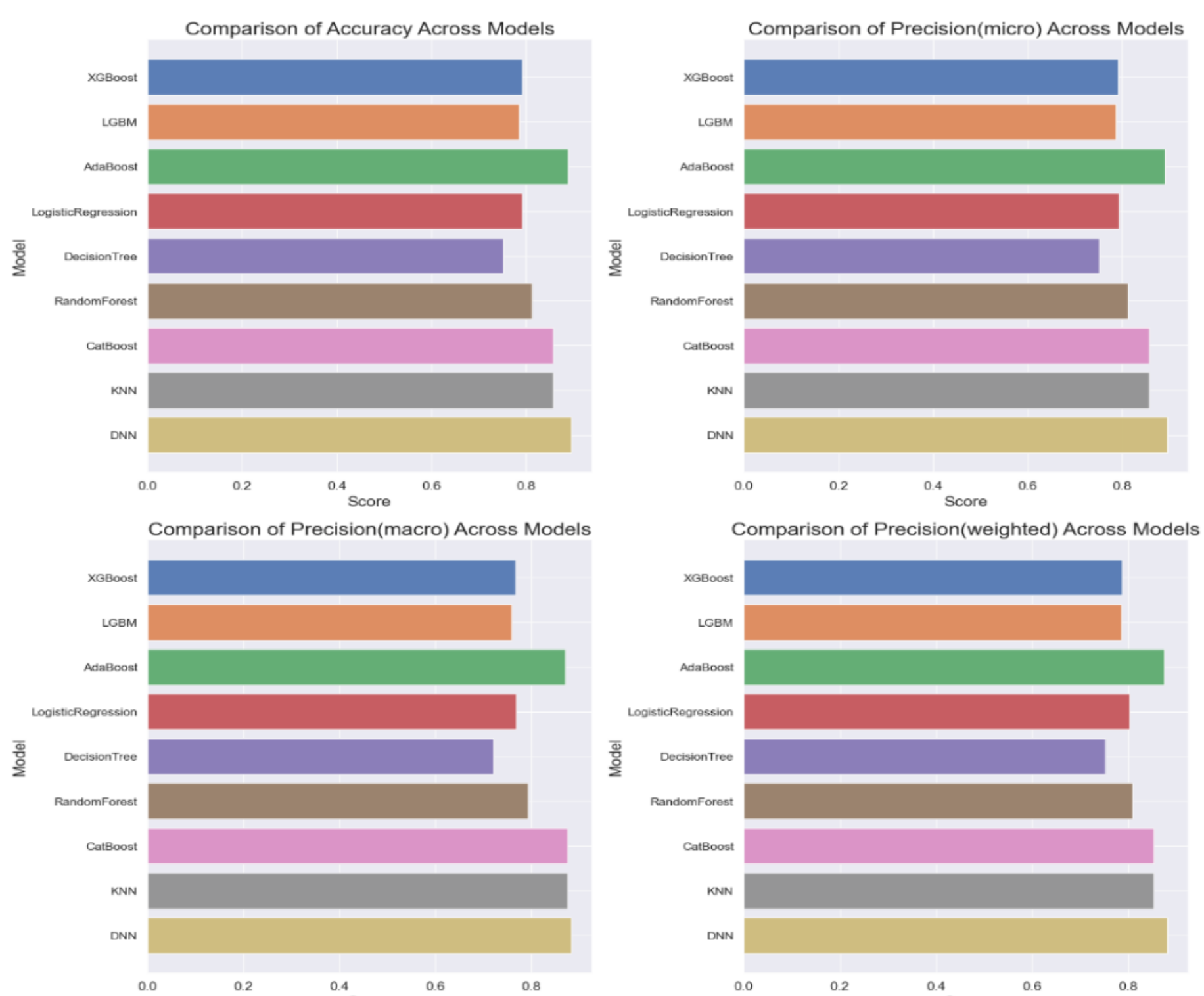

**Figure 14.** Accuracy, Precision and Recall comparison of ML Models.

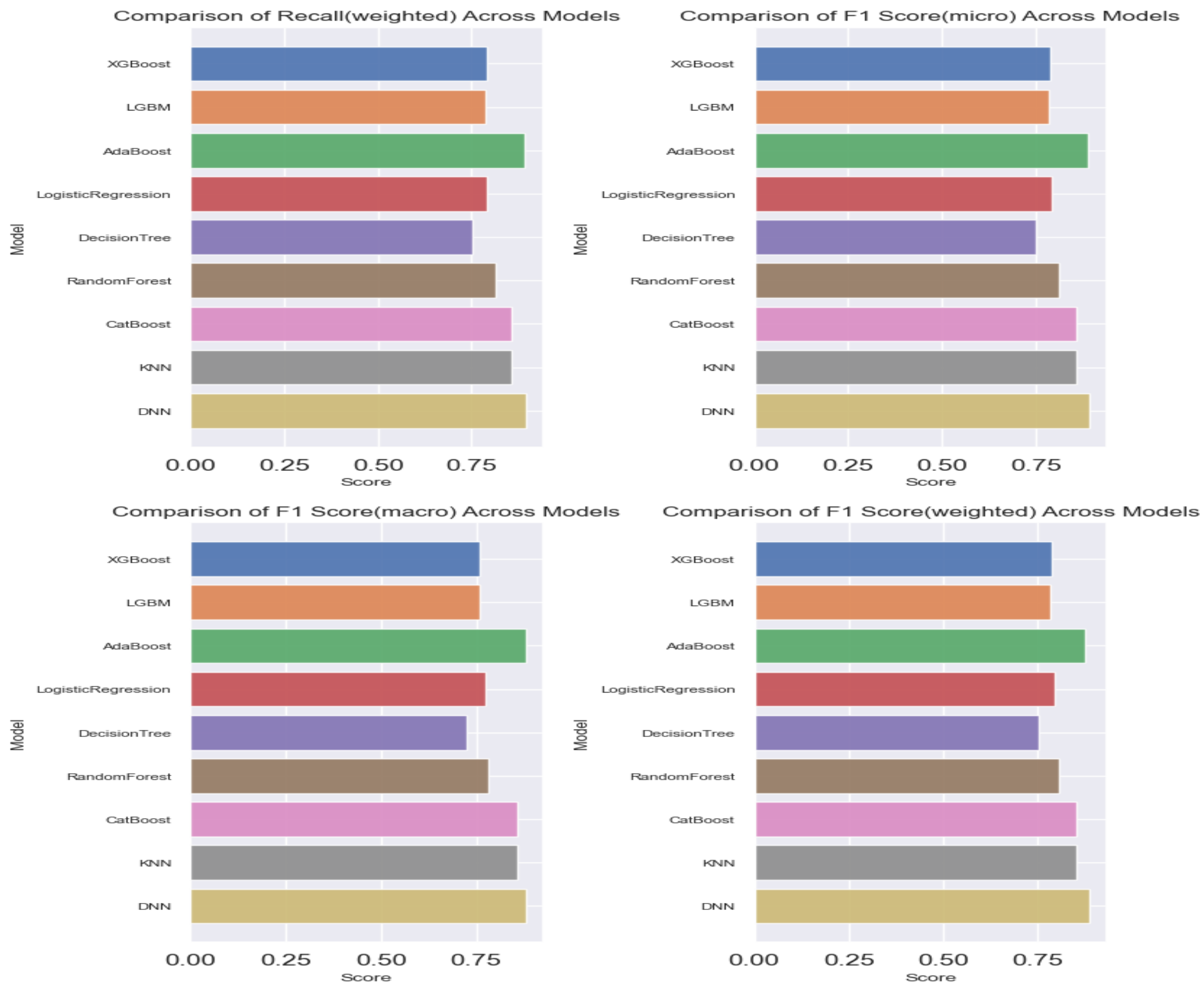

**Figure 15.** Recall and F-1 Score comparison of ML Models.

Moreover, the comparative evaluation across multiple metrics enables researchers to identify trade-offs between different performance criteria and select the most suitable model for specific application scenarios. For instance, while one model may excel in accuracy, it may exhibit lower precision or recall, necessitating careful consideration of the trade-offs involved. Overall, the comprehensive comparisons provided in Figure 13, Figure 14, and Figure 15 offer valuable insights into the relative performance of different ML models in COVID-19 detection. By leveraging these insights, researchers can make informed decisions regarding model selection and optimization strategies, ultimately advancing the development of more accurate and reliable diagnostic tools for combating the ongoing pandemic.

## V. Conclusion

In conclusion, our study underscores the profound potential of machine learning (ML) models in the early detection of COVID-19, offering promising avenues for bolstering pandemic response efforts. Through the meticulous evaluation of various ML algorithms, including DNN, Voting, Bagging, and SVC, among others, we have elucidated their efficacy in predicting COVID-19 infection probability. Our findings highlight the pivotal role of ML techniques in augmenting the efficiency and accuracy of COVID-19 detection, particularly amidst the ongoing challenges posed by the global pandemic. The identification of DNN and Voting models as standout performers, achieving a commendable accuracy of 89%, underscores their utility in facilitating timely identification and containment of the virus. Furthermore, our study emphasizes the critical importance of monitoring and mitigating overfitting in ML models to ensure robust and reliable performance across diverse algorithms. By addressing this challenge, we enhance the credibility and applicability of ML-based COVID-19 detection systems, thereby bolstering confidence in their efficacy in real-world deployment scenarios. Looking ahead, continued research and development efforts in ML-based approaches are imperative for further refining the accuracy, reliability, and scalability of COVID-19 detection methods. By harnessing the power of ML technologies and fostering interdisciplinary collaborations, we can pave the way for more effective pandemic management strategies and mitigate the impact of COVID-19 on a global scale. This underscores the transformative potential of ML-driven innovations in shaping the future of public health response and resilience in the face of infectious disease outbreaks.

## Declarations

### Conflict of Interest

There are no conflicts of interest to disclose.

### Competing Interests

Not applicable.

### Funding Information

No funding.

### Author Contributions

All authors contributed to programming, data collection, writing, and validation.

### Data Availability Statement

Data are accessible upon request by contacting the corresponding author.

### Research Involving Human and/or Animals

Not applicable.

### Informed Consent

Not applicable.